\RequirePackage[final]{graphicx}
\documentclass[12pt,onecolumn,draft,journal]{IEEEtran}
\usepackage{times}
\usepackage{caption}
\usepackage{epsfig}
\usepackage{amsmath}
\usepackage{amssymb}
\usepackage{stfloats}
\usepackage{subcaption}
\usepackage{flafter}
\usepackage[section]{placeins}
\usepackage{wrapfig}

\usepackage{atbegshi}
\AtBeginDocument{\AtBeginShipoutNext{\AtBeginShipoutDiscard}}

\begin{document}

\title{Channel-by-Channel Demosaicking Networks with Embedded Spectral Correlation}
\author{Yan~Niu,~
        Jihong~Ouyang,~
       }
\thanks{Y. Niu and J. Ouyang are with Jilin University e-mail: niuyan@jlu.edu.cn.}


\thispagestyle{empty}
\maketitle
\begin{abstract}
   Demosaicking is standardly the first step in today's Image Signal Processing (ISP) pipeline of digital cameras. It reconstructs image RGB values from the spatially and spectrally sparse Color Filter Array (CFA) samples, which are the original raw data digitized from electrical signals. High quality and low cost demosaicking is not only necessary for photography, but also fundamental for many machine vision tasks. This paper proposes an accurate and fast demosaicking model based on Convolutional Neural Networks (CNN) for the Bayer CFA, which is the most popular color filter arrangement adopted by digital camera manufacturers. Observing that each channel has different estimation complexity, we reconstruct each channel by an individual sub-network. Moreover, instead of directly estimating the red and blue values, our model infers the green-red and green-blue color difference. This strategy allows recovering the most complex channel by a light weight network. Although the total size of our model is significantly smaller than the state of the art demosaicking networks, it achieves substantially higher performance in both demosaicking quality and computational cost, as validated by extensive experiments. Source code will be released along with paper publication.

\end{abstract}

\IEEEpeerreviewmaketitle

\section{Introduction}

\begin{figure}[h]
	\centering
	
		\includegraphics[scale=0.25]{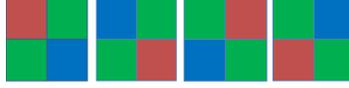}
	\caption{The four layouts of the Bayer pattern CFA.}
	\label{fig:Bayer CFA}
\end{figure}

Current digital cameras mostly use Color Filter Arrays (CFAs), which transduce the light rays deposited at each pixel to one of the RGB intensity values. Consequently, CFA cameras rely on a demosaicking procedure to restore the full RGB channels. The majority of existing methods are designed for the Bayer CFA, which is the most popular sampling pattern in the camera industry. Fig. \ref{fig:Bayer CFA} shows the four layouts: in each $2\times 2$ block of pixels, the diagonal or anti-diagonal positions have their green intensity components captured, while the other two positions have their red and blue intensity components captured. As CFA sampling is not only spatially but also spectrally sparse, it is considered more difficult than $2\times$ image resizing, as noted by Szeliski \cite{szeliski2010}. 

Beside being essential for imaging quality in photography, demosaicking accuracy also significantly affects the performance of subsequent machine vision tasks, as verified by Buckler-Jayasuriya-Sampson \cite{Buckler17}. For higher accuracy, state-of-the-art demosaicking methods adopt Convolutional Neural Networks (CNN) models and tend to go deeper or wider, incurring larger storage space, more frequently memory reading-writing and involving more FLoating Point Operations (FLOPs). However, either consumer-level digital camera photography or real-time machine vision applications make little allowance for computation latency, hence light weight and high quality are of equal importance to demosaicking. 

In this paper, we design a convolutional network composed of three independent light weight sub-networks for Bayer pattern CFA demosaicking. The total number of parameters and FLOPs of the sub-networks is still far fewer than the current top performing demosaicking networks. An obvious reason for using parallel networks is to speed up. The off-line hyper-parameter searching and training of the three networks can be conducted simultaneously on multiple GPUs, which are affordable today; while parallel computation suits for the Field Programmable Gate Arrays (FPGA) implementation of ISP. 

Nevertheless, the most important motivation for using three independent sub-networks is that the RGB color channels have different demosaicking complexity. Particularly for Bayer CFA, the green samples are twice as dense as the other two channels, thus it is clearly the easiest to reconstruct. Nevertheless, even though the red and blue channels are sampled symmetrically, the sensing units of the two channels have different sensitivity, due to the sensor manufacturing. This fact has been rarely considered in previous de-Bayer networks, which generally treat the red and blue channels (otherwise all three channels) in the same fashion. As a consequence, the easiest channel might be over-parameterized or the most difficult channel might be under-parameterized, wasting computation resource or foregoing the accuracy. Therefore, we model the three channels separately.

We transform the red and blue channel reconstruction to green-red and green-blue difference plane reconstruction. It has been verified by Cok \cite{Cok87} that the color difference plane is smoother than the color channels, as the high frequency components of the three channels are mostly consistent. This fact has been widely exploited in hand-crafted interpolation-based demosaicking schemes. For example, Hamilton-Adams used bilinear interpolation to recover the green-red and green-blue difference planes for Bayer demosaicking \cite{HA96}. Albeit being simple, it outperforms many demosaicking algorithms that rely on sophisticated edge detection \cite{Niu2019}.  Our work thus utilizes this technique to reduce the model complexity. By replacing the two color channels with their difference with the green channel, the cross-channel regularization is also naturally embedded to the networks, therefore improving demosaicking quality. 

The proposed demosaicking network is evaluated extensively on benchmark datasets and compared to other demosaicking convolutional neural networks. The experiments show that our algorithm surpasses state-of-the-art networks, while the total trainable parameters of the three networks are about $70\%$ fewer.

The rest of the paper is organized as follows. Section \ref{Sec:Related} reviews the related demosaicking methods. We then introduce the overall workflow of our demosaicking mechanism, as well as the architecture design and training of each individual network in Section \ref{Sec:Proposed}. Section IV evaluates the numerical and visual performance of the proposed algorithm. Section \ref{Sec:Experiments} concludes our work.

\section{Related Works}
\label{Sec:Related}
Demosaicking is challenged by the ``false color'' and ``zippering'' artifacts at image details. To address these problems, numerous demosaicking methods have been proposed, based on explicit interpolations or CNNs.

The interpolation-based methods estimate missing values by a weighted combination of their inter and intra-channel neighbouring pixels, where the weights coming from edge detection. A myriad of mathematical tools have been applied to formulate the inter and intra-channel regularization, for example, Non-local Diffusion \cite{LDINAT11}, Learned Random Fields \cite{MSR2014}, Compressive Sensing \cite{LDSR14}, Dictionary Learning \cite{DDR16}. Iteratively refining one channel by the intermediate demosaicking values can significantly improve the accuracy \cite{MLRI16}, but the iterative computation hinders their application in real practice \cite{Niu2019}. The High Quality Linear Interpolation (HQLI) by Malvar-He-Culter is probably the only linear (except the basic Bi-linear interpolation) and the fastest interpolation algorithm in the literature \cite{HQLI04}. We employ HQLI to initialize the input to our networks. The Hamilton-Adams algorithm is another very simple yet effective method \cite{HA96}, which inspires us to transform the red and blue recovery to color-difference inference. 

The CNN based methods learn the local image spatial and spectral correlation from labeled data, thereby defining a mapping function from CFA samples to RGB images. Wang applied a 3-layer neural network to demosaicking, learning the weights and bias parameters by a combination of supervised and unsupervised learning \cite{MNN14}. Tan et al. designed a two-stage convolutional network of 11 convolution layers, taking the bilinearly demosaicked images as the initial input \cite{DRL17}. Its accuracy and efficiency outperform concurrent interpolation demosaicking methods. Gharbi et al. proposed a network of 15 convolution layers for jointly demosaicking and denoising (referred to as JDD) \cite{DJ16}. It re-arranges the CFA image to four input channels. Compared to taking an initially demosaicked image as input, this re-arrangement reduces the FLOPs of the network. Kokkinos-Lefkimmiatis investigated jointly demosaicking and denoising in an iterative Expectation-Maximization framework. In each of the 10 iterations, the intermediate demosaicking is refined and then denoised by a 5-layer pretrained denoising network \cite{CAS18}. Cui-Jin-Steinbach extended the two-stage demosaicking network to a three-stage network with 20 convolution layers, each of which has 128 filters. It is one of the top-performing networks, but the computational cost is also high. Recently, demosaicking networks for burst photography, which takes multiple images at one shutter click, have also been developed \cite{BurstCVPR19} \cite{BurstICCV19}. 

Existing demosaicking networks are mainly designed for the Bayer CFA, which samples the green channel more densely than the other two channels. Most of de-Bayer networks treat the three-channels without distinguishing, except that the two-stage network and three-stage network reconstruct the green channel separately from the other channels. In contrast, we handle each channel by a sub-network individually designed. By this strategy, our blue channel reconstruction accuracy significantly outperforms the current top accuracy. 

The use of multi-networks has also been discussed for demosaicking. The parallel networks by Tan-Chen-Hua address the image flat regions, edges and textures separately \cite{MDFCN18}. Yamaguchi-Ikehara used four independent networks, each of which consists of three independent sub-networks, mapping the CFA samples to the chrominance images \cite{ICASSP19}. The design of our network is fundamentally different from these works and has significantly lower cost.
  
Huang et al. proposed a light weight network (referred to as LW) \cite{huang2018}, by applying weight-reducing techniques such as bottleneck layer and aggregated convolution transformations. This is so far the smallest network in the literature of CNN based demosaicking, having the fewest parameters and FLOPs. However, as pointed out by \cite{Shuffle18}, fewer FLOPs does not necessarily mean faster computation speed. Particularly, as the LW network contains 36 convolution layers, the total time cost on memory reading-writing of all layers is not negligible. To save memory storage and input-output time, We restrict the architecture searching space by depth and number of parameters.  
\section {Light Weight and High Quality Demosaicking Networks}
\label{Sec:Proposed}

\subsection{The Overall Workflow}
\label{subsec:WorkFlow}
\begin{figure*}[t]
	\centering
		\includegraphics[scale=1.2]{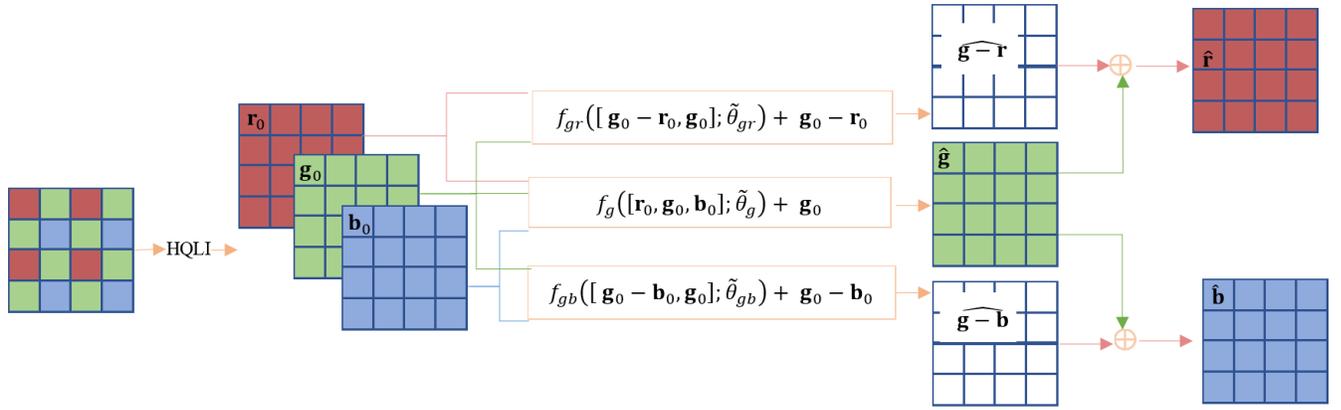}
	\caption{The overall work-flow of the proposed CNN-based demosaicking process. $\tilde{\theta}_{g}$, $\tilde{\theta}_{gr}$ and $\tilde{\theta}_{gb}$ are the network parameters obtained through training; symbol $\oplus$ stands for the general addition operations. Particularly in the algorithm, they perform the subtraction computations.}
	\label{fig:workflow}
\end{figure*}

Fig.\ref{fig:workflow} outlines the overall work-flow of the proposed method. 

Many CNN-based demosaicking methods initialize the inputs by a fast interpolation first (e.g., \cite{DRL17}; \cite{MDFCN18}). The initialization predicts the missing values at low cost. Especially in flat image regions, which are the main part of a real-world image, the prediction is close to the true values. We follow this routine and initialize the network inputs by the HQLI algorithm \cite{HQLI04}. Briefly, HQLI enhances the bilinear reconstruction of one channel by the details (Laplacians) of another channel. Still, it suffers the ``false color'' and ``zippering'' artifacts, to be corrected by the subsequent networks.

We define a Bayer CFA sampled image patch by $\mathbf{M}$; Its original red, green and blue channels by $\mathbf{r}$, $\mathbf{g}$ and $\mathbf{b}$. Notations $\mathcal{R}$, $\mathcal{G}$ and $\mathcal{B}$ stand for the sets of pixels whose red, green and blue components are originally available. In general, a demosaicking network models a function $f$, mapping from the HQLI initialization $[\mathbf{r}_{0}, \mathbf{g}_{0}, \mathbf{b}_{0}]$ to the true image $[\mathbf{r}, \mathbf{g}, \mathbf{b}]$. We disentangle $f$ to three functions $f_{gr}$, $f_{g}$ and $f_{gb}$, restoring the green channel, the green-red difference plane and the green-blue difference plane:  
\begin{align}
	\begin{split}
    f_{gr}: &[\mathbf{g}_{0}-\mathbf{r}_{0},\mathbf{g}_{0},\mathbf{b}_{0}] \mapsto \mathbf{g}-\mathbf{r}, \nonumber\\
		f_{g}:  &[\mathbf{r}_{0},\mathbf{g}_{0},\mathbf{b}_{0}] \mapsto \mathbf{g}, \nonumber\\  
		f_{gb}: &[\mathbf{r}_{0},\mathbf{g}_{0},\mathbf{g}_{0}-\mathbf{b}_{0}] \mapsto \mathbf{g}-\mathbf{b}.
	\end{split}
\end{align} 
Here symbol $\mapsto$ means ``maps to''. Note that $\mathbf{r}_{0}(\mathcal{R})=\mathbf{r}(\mathcal{R})$, $\mathbf{g}_{0}(\mathcal{G})=\mathbf{g}(\mathcal{G})$ and $\mathbf{b}_{0}(\mathcal{B})=\mathbf{b}(\mathcal{B})$, according to the CFA sampling.

In empirical study, we observe that $\mathbf{b}_0$ shows trivial contribution to the demosaicking performance of $f_{gr}$, and so does $\mathbf{r}_0$ to $f_{gb}$. Furthermore, we employ the residual network \cite{he2016deep}. Thus the mapping functions are simplified to
 \begin{align}
	\begin{split}
    f_{gr}: &[\mathbf{g}_{0}-\mathbf{r}_{0},\mathbf{g}_{0}] \mapsto (\mathbf{g}-\mathbf{r})-(\mathbf{g}_{0}-\mathbf{r}_{0}), \nonumber\\
		f_{g}:  &[\mathbf{r}_{0},\mathbf{g}_{0},\mathbf{b}_{0}] \mapsto \mathbf{g}-\mathbf{g}_{0}, \nonumber\\  
		f_{gb}: &[\mathbf{g}_{0}-\mathbf{b}_{0},\mathbf{g}_{0}] \mapsto (\mathbf{g}-\mathbf{b})-(\mathbf{g}_{0}-\mathbf{b}_{0}).
	\end{split}
\end{align}
We model these mapping functions by convolutional networks with parameters $\theta_{gr}$,$\theta_{g}$ and $\theta_{gb}$, and refer to these networks as $f_{gr}(\quad;\theta_{gr})$, $f_{g}(\quad;\theta_{g})$ and $f_{gb}(\quad;\theta_{gb})$. 

\subsection{Network Design}
\begin{figure*}[h]
	\centering
         \includegraphics[width=\textwidth]{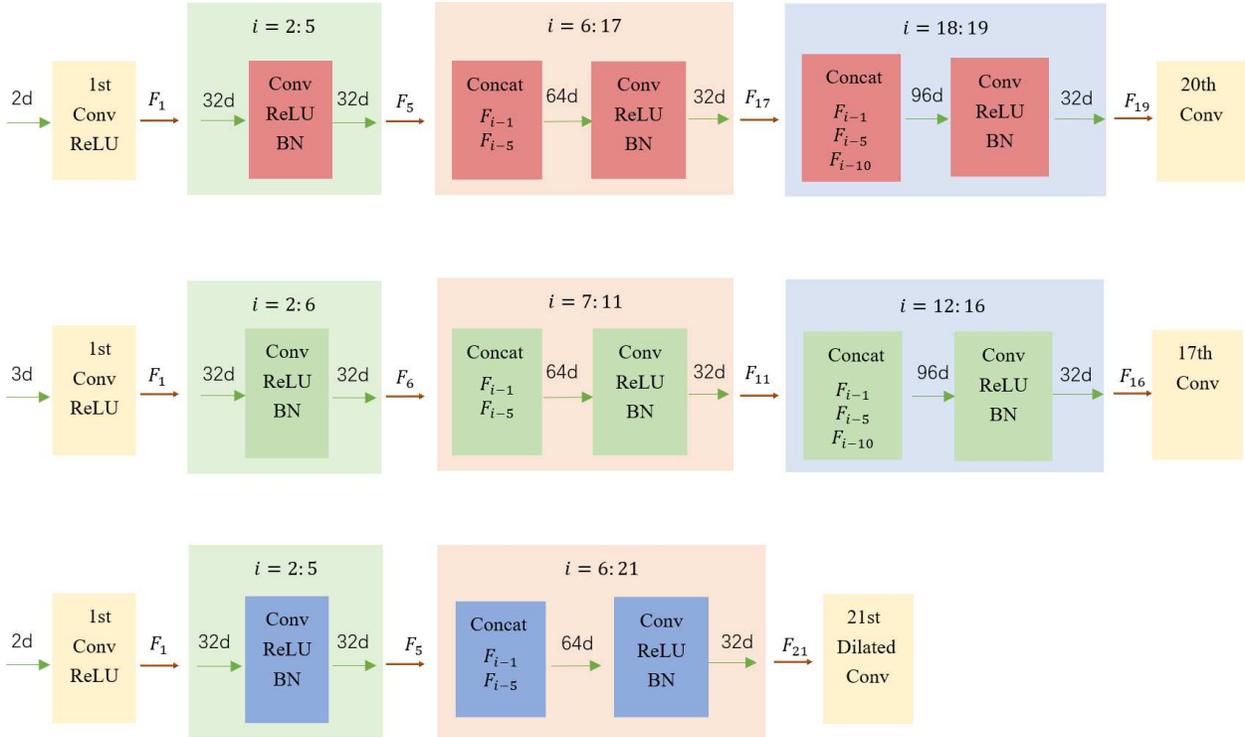}	
	\caption{The network architecture of $f_{gr}$, $f_{g}$ and $f_{gb}$ (from top to bottom), obtained by progressively searching for the proper width, depth and number of parameters. The hidden layers of each network are indexed by $i$. The networks have totally 924,480 trainable parameters and takes 10.14MB memory.}
	\label{fig:architecture}
\end{figure*}

Fig. \ref{fig:architecture} depicts the architecture of our demosaicking networks.

We construct each network by stacking $D$ convolution blocks, including an input layer, $D-2$ hidden layers and a prediction layer. Each convolution uses $K$ filters of spatial support $3\times 3$ (unless otherwise specified), as Cudnn is optimized for $3\times 3$ convolution \cite{Shuffle18}. The input layer contains a convolution and a Rectified Linear Unit (ReLU) activation. It maps the $M\times N\times C$ ($C=3$ for $f_{g}$, $C=2$ for $f_{gr}$ and $f_{gb}$) input tensor to $M\times N\times K$ feature channels. Each feature extraction layer consists of convolution, batch normalization and ReLU activation. The last layer predicts the output from $M\times N\times K$ feature maps by pure convolution. We search for proper values of $D$ and $K$, requiring the number of parameters $P$ of each sub-network to be up-bounded by $6\times 10^5$, which is roughly the size of the light weight JDD network. The network depth $D$ is limited by 40.

Assuming the network has no shortcut connections, then the up-bound for $D$ can be computed by $ d(K) = \min(40,\left\lfloor \frac{6\times 10^5 - 9(C\times K + K)}{K^2}\right\rfloor+2$). 
We then train the networks of hyper-parameters $(d(K),K)$, with $K\in \left\{32,64,128\right\}$, and test them on the validation sets. Let $(d(\tilde{K}),\tilde{K})$ be the hyper-parameters associated with the smallest validation error. For all channels, we observed that $\tilde{K}=32$ performs the best. Thus given $(d(32), 32)$, we progressively remove the hidden layers at a step-size, which is initially set to 5 then 2 along with the shrinking of $d(\tilde{K})$, until the validation error starts increasing. This progress decreases the number of parameters to $\tilde{P}$. Next, we fix the number of parameters and further remove the depth by establishing skip connections, which concatenate the low level and high level features similarly to dense connection but sparsely. In particular, from the 6th layer, we concatenate the feature output of the $(i-1)$-th layer with the feature outputs of the $(i-5)$-th layer, and optionally with the $(i-10)$-th layer. We train the networks by a few epochs and compare their validation errors. The selection obtains $\tilde{D}$. Finally, because the red and blue channel samples are sparser than the green channel, we include more original samples to green-red and green-blue difference recovery, by applying dilated convolution to increase the receptive field without extra parameters. We dilate the $3\times 3$ convolution kernel to $7\times 7$ by inserting zeros between the neighbouring elements.

To train $\theta_{g}$ for $f_{g}(\quad;\theta_{g})$ , we define tensor 
\[\mathbf{X}_{i,g} = [\mathbf{r}_{0}, \mathbf{g}_{0}, \mathbf{b}_{0}]_{i}\]
as the $i$th training observation, and tensor
\[\mathbf{Y}_{i,g} = [\mathbf{g}-\mathbf{g}_{0}]_{i}\]
as its associated label. The desirable $\theta_g$ should have the prediction $f_g(\mathbf{X}_{i,g};\theta_g)$ very close to $\mathbf{Y}_{i,g}$. Our model estimates not only $\mathbf{g}(\mathcal{R})$ and $\mathbf{g}(\mathcal{B})$, but also $\mathbf{g}(\mathcal{G})$, meaning that the true values for half of the prediction are originally available in the input. This is similar to the auto encoder-decoder network, which extracts the underlying structure by reconstructing the inputs \cite{autoencoder14}. Therefore, our green channel recovery network adopts the Mean Squared Error (MSE) loss function of the auto encoder-decoder
\begin{equation}
	e_{g} = \frac{1}{n}\sum_{i=1}^{n}{\left\|f_{g}(\mathbf{X}_{i,g};\theta_g)-\mathbf{Y}_{i,g}\right\|_{2}^{2}}.
\end{equation}
Here $n$ stands for the number of training observations. 

The networks for $f_{gr}(\quad;\theta_{gr})$ and $f_{gb}(\quad;\theta_{gb})$ take the tensors
\begin{align*}
	\begin{split}
	\mathbf{X}_{i,gr} = [\mathbf{g}_{0}-\mathbf{r}_{0}, \mathbf{g}_{0}]_{i}\nonumber \\
	\mathbf{X}_{i,gb} = [\mathbf{g}_{0}-\mathbf{b}_{0}, \mathbf{g}_{0}]_{i}
	\end{split}
\end{align*}
as the $i$th input respectively. Their label tensors are 
\begin{align*}
	\begin{split}
		\mathbf{Y}_{i,gr} &= [(\mathbf{g}-\mathbf{r})-(\mathbf{g}_0-\mathbf{r}_0)]_{i}\nonumber \\
		\mathbf{Y}_{i,gb} &= [(\mathbf{g}-\mathbf{b})-(\mathbf{g}_0-\mathbf{b}_0)]_{i}.
	\end{split}
\end{align*}	

Different from the green channel, the green-red or green-blue difference recovery does not have the auto encoder-decoder nature. Particularly, at each pixel, either $\mathbf{g}_{0}$ or $\mathbf{r}_{0}$ (or both) in $[\mathbf{g}_{0}-\mathbf{r}_{0}, \mathbf{g}_{0}]_{i}$ is generated by estimation. That is, none of the input values is identical to its labels for $f_{gr}(\quad;\theta_{gr})$. The situation for $f_{gb}(\quad;\theta_{gb})$ is the same. For robustness to outliers, we adopt the $p$-norm, where $p=0.9$, as the penalty function for the objective functions. Thus we define 
\begin{equation}
	e_{gr} = \frac{1}{n}\sum_{i=1}^{n}{\left\|f_{gr}(\mathbf{X}_{i,gr};\theta_{gr})-\mathbf{Y}_{i,gr}\right\|_{p}},
\end{equation}
and 
\begin{equation}
	e_{gb} = \frac{1}{n}\sum_{i=1}^{n}{\left\|f_{gb}(\mathbf{X}_{i,gb};\theta_{gb})-\mathbf{Y}_{i,gb}\right\|_{p}}.
\end{equation}

\subsection{Network Training}
\label{subsec:aug}
To simulate the input observations, the training images should be subsampled to CFA images. As the available training images are limited, many previous works augment the training data by flipping and rotating the images. This method works well for classification networks to extract pose-invariant features. However, for regression problems such as demosaicking, the augmentation does not supply much extra information. Some previous works augment extra training images by collecting low resolution images from social webs. The statistical distribution of the training data obtained in this way has different nature from the raw data.  

To train the green recovery network, we perform data augmentation by simply removing each training image's first row and first column respectively, to generate two extra training images. For the other two networks, we additionally take in the images that have both their first column and first row removed. As each sub-network is light weighted, although our training dataset (the Waterloo Exploration Database \cite{Waterloo17}) contains only 4744 images, the augmentation is sufficient to avoid over-fitting.

We randomly shuffle the order of the images, and then partition each image to $50\times 50$ patches. Patches from the first $95\%$ images are used for training, we then discard the next 1792 consecutive patches, and use the rest patches for validation. This ensures that the training and validation patches are from different images, thus no validation patch nearly duplicates a training patch. Each training batch contains 128 input-label pairs. The cost functions $e_{g}$, $e_{gr}$ and $e_{gb}$ are minimized by the Adaptive Moment Estimation (ADAM) optimizer \cite{kingma2015} independently. For all training procedures, the initial learning rate is set to 0.005, then is halved every $5$ epochs until reaching $\frac{0.005}{64}$. Our offline training is implemented by the MatConvNet toolbox \cite{vedaldi15matconvnet} in Matlab, and run on a couple of 11GB Nvidia GTX 2080Ti and a RTX 2070Super GPUs, each for one model. Each epoch takes about 150 minutes. All models show clear convergence around 30 epochs, at which point the validation error varies trivially, therefore the parameters obtained by the 30th epoch are used for online testing.  

The training obtains $\tilde{\theta}_{g}$, $\tilde{\theta}_{gr}$ and $\tilde{\theta}_{gb}$, which are used for online demosaicking. The green network has 3.04 MB memory footprint and $277,632$ trainable parameters; the green-red network is of 3.45 MB memory and $314,208$ parameters; the green-blue network is of 3.65 MB and $332,640$ parameters. Compared to existing demosaicking networks, each of our sub-networks is small, and the total number of parameters is about $31.7\%$ of the 3-Stage network.

\subsection{Demosaicking} 
$\mathbf{g}$, $\mathbf{g}-\mathbf{r}$ and $\mathbf{g}-\mathbf{b}$ of a mosaicked image $\mathbf{M}$ are finally estimated by:
\begin{align}
	\mathbf{\hat{g}} & =  f_{g}([\mathbf{r}_{0},\mathbf{g}_{0},\mathbf{b}_{0}];\tilde{\theta}_g) + \mathbf{g}_{0}\nonumber \\
	\mathbf{\widehat{g-r}} & =  f_{gr}([\mathbf{g}_{0}-\mathbf{r}_{0},\mathbf{g}_{0}];\tilde{\theta}_{gr}) + \mathbf{g}_{0}-\mathbf{r}_{0}\nonumber \\
	\mathbf{\widehat{g-b}} & =  f_{gb}([\mathbf{g}_{0}-\mathbf{b}_{0},\mathbf{g}_{0}];\tilde{\theta}_{gb}) + \mathbf{g}_{0}-\mathbf{b}_{0}.
\end{align}
This gives
\begin{align}
 \begin{split}
  \hat{\mathbf{g}}(\mathcal{G}) & = \mathbf{g(\mathcal{G})}\nonumber \\  
	\hat{\mathbf{r}}(\mathcal{R}) & = \mathbf{r(\mathcal{R})} \quad \hat{\mathbf{r}}(\mathcal{G}\cup\mathcal{B}) = (\mathbf{\hat{g}}-\widehat{\mathbf{g}-\mathbf{r}})(\mathcal{G}\cup\mathcal{B})\nonumber \\
	\hat{\mathbf{b}}(\mathcal{B}) & = \mathbf{b(\mathcal{B})} \quad \hat{\mathbf{b}}(\mathcal{G}\cup\mathcal{R}) = (\mathbf{\hat{g}}-\widehat{\mathbf{g}-\mathbf{b}})(\mathcal{G}\cup\mathcal{R}),
 \end{split}
\end{align}
Note that some estimated values may be out of the range $[0,255]$. Such values are clipped to $0$ or $255$.

\section{Experimental Results}
\label{Sec:Experiments}
The online demosaicking is tested on an Intel i7-8700 CPU clocked at 3.2GHz with 64GB RAM. The demosaicking literature commonly evaluate the algorithms on benchmark datasets Kodak \cite{Kodak} (obtained by color scanning) and McM \cite{LDINAT11} (cropped from a high resolution image). A few works also adopt benchmarks MSR \cite{MSR2014}, designed to test demosaicking in the linear color space, and MIT Hard Patch including Moire and Hdrvdp \cite{JDD2018}, designed to test how well an algorithm overcomes the false color and zippering artifacts respectively. 
Our experiments are conducted on these datasets.

\subsection{Numerical Evaluation}
Table \ref{tab:CNNComparison} lists the average Peak Signal to Noise Ratio (PSNR) values of the proposed and state-of-the-art demosaicking convolutional neural networks on the test benchmarks\footnote{The results reported for MSR are obtained on MSR's noise-free linear Bayer Panasonic test dataset}. The PSNRs of the 2-Stage and the 3-Stage networks are obtained by running their released code; results of other methods are as reported in \cite{JDD2018} and \cite{CAS18}. It should be mentioned that, the tested accuracy varies drastically with the training dataset employed. As pointed out by Kokkinos-Lefkimmiatis \cite{CAS18}, training their Cascaded network by the MIT-training set can increase the testing accuracy on MIT-Hdrvdp and MIT-Moire by 5.3 dB and 7.3 dB over its MSR-training counterparts. In the experiments, we do not fine tune our networks. Table \ref{tab:CNNComparison} also lists the number of trainable parameters and Flops involved in each model. Since the input layer and prediction layer of all these networks are similar, and they contain a trivial number of parameters, we only count the parameters for hidden layer convolutions. The hidden layer FLOPs are computed assuming that the CFA image is $100\times 100$. 

On the Kodak and MIT-vdp datasets, our network achieves the top accuracy; on McM and MSR, the PSNRs of our network rank the second-best. The PSNR comparison shows that our networks, trained on the WED dataset, generalizes well to images of various distributions. The PSNRs of the proposed networks on McM and MSR are 0.18 dB and 0.6 dB lower than the top scores achieved by the Cascaded network, whose FLOPs roughly doubles our model. The 3-Stage network has accuracy comparable to ours, but its FLOPs and number of parameters are three times as ours. Moreover, our network is considerably smaller than the methods Multiple and Parallel, which also employ multiple networks.

\begin{table}[h]
  \centering
    \begin{tabular}{c|c|c|c|c|c|c|c}
		\hline
		\hline
		{} & {Kodak} & {McM} & {MSR} & {MIT-Moire} & {MIT-vdp} &{Paras ($\times 10^5$)} & {Flops ($\times 10^9$)}\\
		\hline
		{JDD \cite{DJ16}} & 41.20 & 39.50 &{41.6} &{\textbf{37.0}} &{34.3} & $5.59$ & $1.39$\\
		\hline
		{Cascade \cite{CAS18}}  & 41.50 & \textbf{39.70} &{\textbf{42.6}} &{\textbf{37.0}} &{34.5}& $\textbf{1.84}$ & $18.4$ \\
		\hline
		{Parallel \cite{ICASSP19}}  & 41.88 & 38.18 &{-} &{-} &{-} & $206$ & $125$\\
		\hline
		{Multiple \cite{MDFCN18}} & 42.04 & 37.62 &{-} &{-} &{-} & $43.5$ & $43.5$\\
		\hline	
		{2-Stage \cite{DRL17}} & 42.04 & 38.98 &{-} &{33.0} &{34.3} & $2.29$ & $2.29$ \\
		\hline
		{3-Stage \cite{3Stage18}} & 42.39 & 39.39 &{-} &{33.3} &{34.7} & $29.49$ & $29.4$\\	
		\hline
		{Light \cite{huang2018}} & 42.60 & 39.21 &{-} &{-} &{-} & $2.26$ & $ \textbf{0.56}$ \\
		\hline
		{Proposed} & {\textbf{42.70}} & {39.52} &{42.0} &{33.5} &{\textbf{35.1}} & $9.25$ & $9.25$\\
		\hline
		\hline
		\end{tabular}
		\vspace{1ex}
		\caption{The PSNRs on benchmarks, number of hidden layer parameters, and Flops of the proposed network and state-of-the-art de-Bayer networks. The Flops are computed assuming image size $100\times 100$.}
		\label{tab:CNNComparison}
\end{table}
 
\begin{figure*}[h!]
	\centering
	\begin{subfigure}[b]{0.15\textwidth}
         \centering
         \includegraphics[scale=1.5]{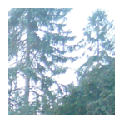}
     \end{subfigure}
		~
		\begin{subfigure}[b]{0.15\textwidth}
         \centering
         \includegraphics[scale=1.5]{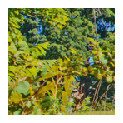}
     \end{subfigure}
		~
		\begin{subfigure}[b]{0.15\textwidth}
         \centering
         \includegraphics[scale=1.5]{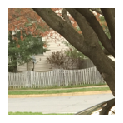}
     \end{subfigure}
	~
		\begin{subfigure}[b]{0.15\textwidth}
         \centering
         \includegraphics[scale=1.5]{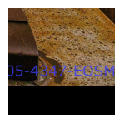}
     \end{subfigure}
	~
		\begin{subfigure}[b]{0.15\textwidth}
         \centering
         \includegraphics[scale=1.5]{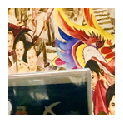}
     \end{subfigure}
\\
\vspace{0.5cm}
		\begin{subfigure}[b]{0.15\textwidth}
         \centering
         \includegraphics[scale=1.5]{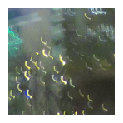}
     \end{subfigure}
		~
		\begin{subfigure}[b]{0.15\textwidth}
         \centering
         \includegraphics[scale=1.5]{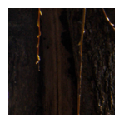}
     \end{subfigure}
		~
		\begin{subfigure}[b]{0.15\textwidth}
         \centering
         \includegraphics[scale=1.5]{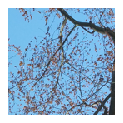}
     \end{subfigure}
		~
		\begin{subfigure}[b]{0.15\textwidth}
         \centering
         \includegraphics[scale=1.5]{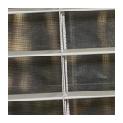}
     \end{subfigure}
		~
		\begin{subfigure}[b]{0.15\textwidth}
         \centering
         \includegraphics[scale=1.5]{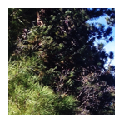}
     \end{subfigure}
\caption{The first 10 images of the MIT-Moire test dataset.}
	\label{fig:MIT-Moire-Test}
\end{figure*}

To inspect the performance of each sub-networks, we compare its PSNRs on each channel with the 2-Stage and 3-Stage networks, by running their released Matlab source code on Kodak and McM, as well as the first ten test images of MIT-Moire (Fig.\ref{fig:MIT-Moire-Test}). As shown by Table \ref{tab:codeComparison}, compared to the 3-Stage network, our network's red PSNR is 0.38 dB higher and the blue PSNR is 0.26 dB higher on Kodak. This noticeable improvement is due to modeling the recovery of the two channels separately. 

\begin{table*}[htb]
  \centering
    \begin{tabular}{c|c|c|c|c|c|c|c|c|c|c|c}
		\hline
		\hline
		\multicolumn{3}{c|}{} & \multicolumn{3}{c|}{Kodak} & \multicolumn{3}{c|}{McM} & \multicolumn{3}{c}{MIT-Moire (10 images)}\\
		\cline{4-6}\cline{7-9} \cline{10-12}
		\multicolumn{3}{c|}{} & \multicolumn{1}{c|}{R} & \multicolumn{1}{c|}{G} & \multicolumn{1}{c|}{B} 
		                      & \multicolumn{1}{c|}{R} & \multicolumn{1}{c|}{G} & \multicolumn{1}{c|}{B} 
													& \multicolumn{1}{c|}{R} & \multicolumn{1}{c|}{G} & \multicolumn{1}{c}{B}\\
		\hline
		\hline
		\multicolumn{3}{c|}{2-Stage \cite{DRL17}} & 41.38 & 44.85 & 41.04  & 39.14 & 42.10 & 37.31 & {34.39} & {37.34} & {33.13}\\
		\hline
		\multicolumn{3}{c|}{3-Stage} & 42.07 & 45.18 & 41.09 & 39.60 & 42.60 & 37.68 & {34.57} & {37.59} & {33.32}\\
	  \hline
		\multicolumn{3}{c|}{Proposed} & \textbf{42.45} & \textbf{45.49} & \textbf{41.35} & \textbf{39.68} &\textbf{42.75} & \textbf{37.86} & \textbf{34.82} & \textbf{37.87} & \textbf{33.72}\\
		\hline
		\hline
		\end{tabular}
		\vspace{1ex}
		\caption{Demosaicking accuracy measured by PSNR on each channel of the proposed method and state of the art demosaicking methods that have Matlab code released. Bold numbers highlight the best preformance in each column.}
		\label{tab:codeComparison}
\end{table*}

\subsection{Visual Performance}
\begin{figure*}[h]
	\centering
		\begin{subfigure}[b]{0.455\textwidth}
         \centering
         \includegraphics[width=1.0\textwidth]{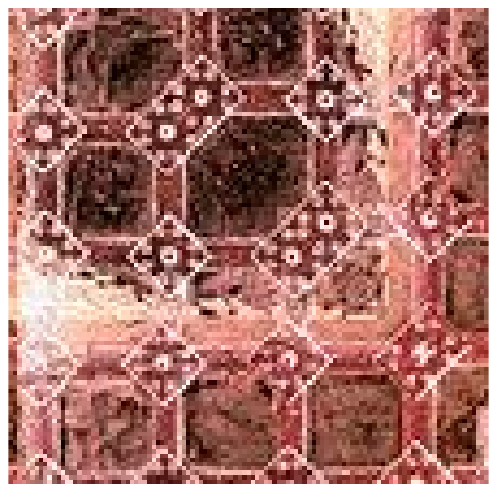}
         \caption{Original}
         \label{subfig:oriTexture}
     \end{subfigure}
		~
		\begin{subfigure}[b]{0.455\textwidth}
         \centering
         \includegraphics[width=1.0\textwidth]{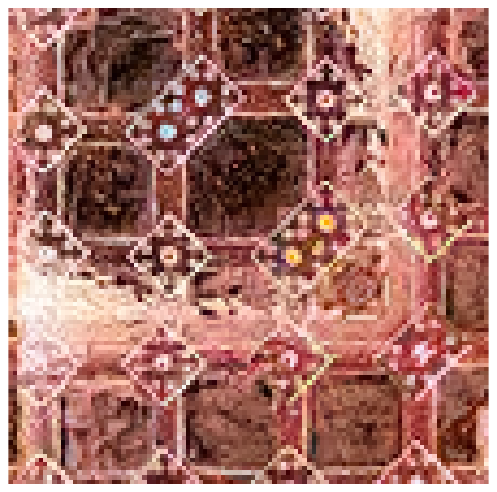}
         \caption{2-Stage}
         \label{subfig:2StageTexture}
     \end{subfigure}
\\
		\begin{subfigure}[b]{0.455\textwidth}
         \centering
         \includegraphics[width=1.0\textwidth]{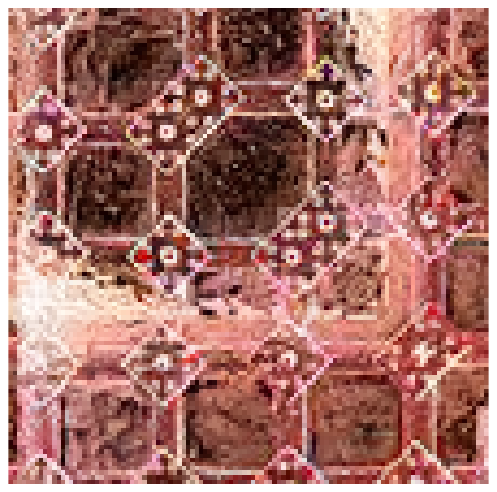}
         \caption{3-Stage}
         \label{subfig:3StageTexture}
     \end{subfigure}
	~
		\begin{subfigure}[b]{0.455\textwidth}
         \centering
         \includegraphics[width=1.0\textwidth]{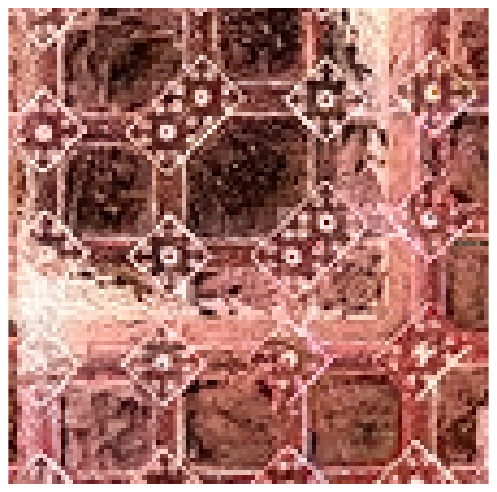}
         \caption{Proposed}
         \label{subfig:NewTexture}
     \end{subfigure}	
 \caption{Visual results on test image google\_lined\_167 from SuperTex136 by the 2-Stage network, 3-Stage network and the proposed network. Better viewed digitally.}
 \label{fig:texture}
\end{figure*}

In addition to the quantitative evaluation, we present the visual performance of our model on challenging texture images. Fig. \ref{subfig:2StageTexture} - \ref{subfig:NewTexture} show an example on image google\_lined\_167 taken from the texture dataset SuperTex136 \cite{SuperTex}, which is frequently used to test Image Super-Resolution Reconstruction algorithms. In this example, the 2-Stage networks mis-estimates the white circles as filled blue or yellow circles; the 3-Stage network mis-estimates the corners of the whilte lines as pink or red. In contrast, our network reconstructs these areas with high fidelity. 
   
\begin{figure*}[h]
 \centering
		\begin{subfigure}[b]{0.455\textwidth}
         \centering
         \includegraphics[width=1.0\textwidth]{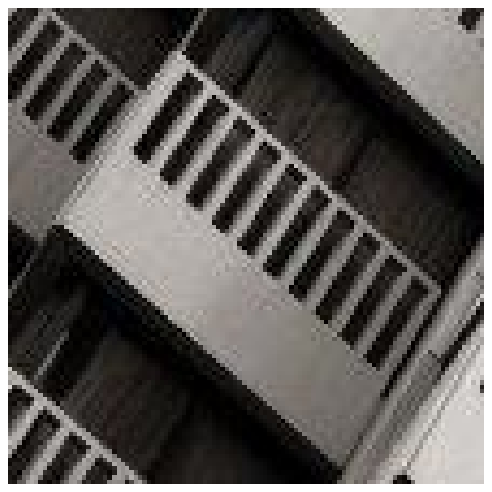}
         \caption{Original}
         \label{subfig:oriUrban}
     \end{subfigure}
~		
		\begin{subfigure}[b]{0.455\textwidth}
         \centering
         \includegraphics[width=1.0\textwidth]{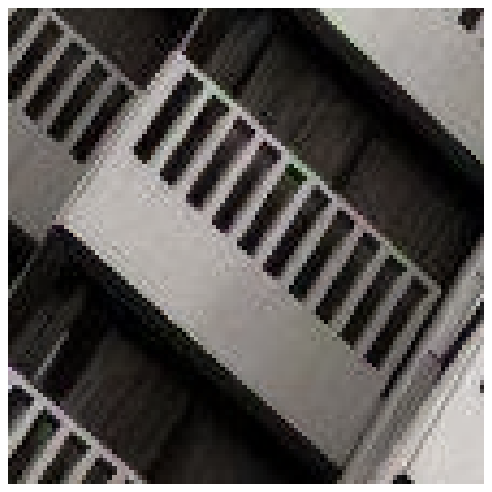}
         \caption{2-Stage}
         \label{subfig:2StageUrban}
     \end{subfigure}
\\
	  \begin{subfigure}[b]{0.455\textwidth}
         \centering
         \includegraphics[width=1.0\textwidth]{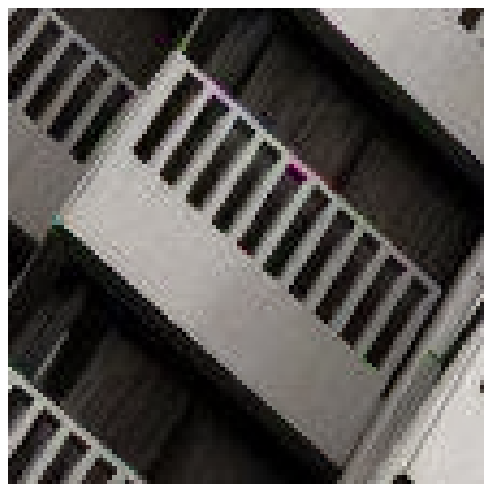}
         \caption{3-Stage}
         \label{subfig:3StageUrban}
		\end{subfigure}
	~
		\begin{subfigure}[b]{0.455\textwidth}
         \centering
         \includegraphics[width=1.0\textwidth]{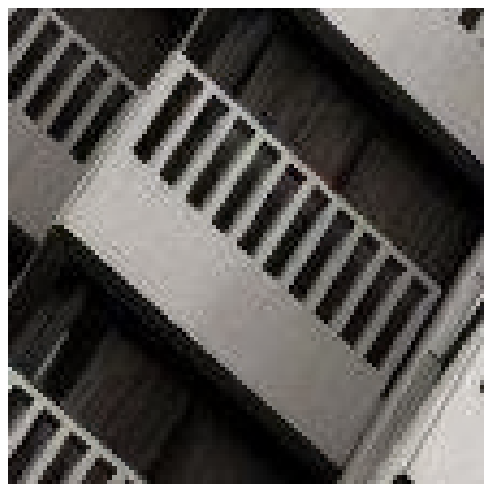}
         \caption{Proposed}
         \label{subfig:NewUrban}
     \end{subfigure}
	\caption{Visual results on test image google\_urban\_4 from SuperTex136 by the 2-Stage network, 3-Stage network and the proposed network. Better viewed digitally.}
	\label{fig:urban}
\end{figure*}

Fig. \ref{subfig:2StageUrban} - \ref{subfig:NewUrban} shows another qualitative comparison example on the Image google\_urban\_4 taken from SuperTex136. The 2-Stage and 3-stage networks suffer obvious false colors at edges, whereas our recovery is sharp and faithful to the ground truth.

\begin{figure*}[h]
	\centering
	\begin{subfigure}[b]{0.455\textwidth}
         \centering
         \includegraphics[width=1.0\textwidth]{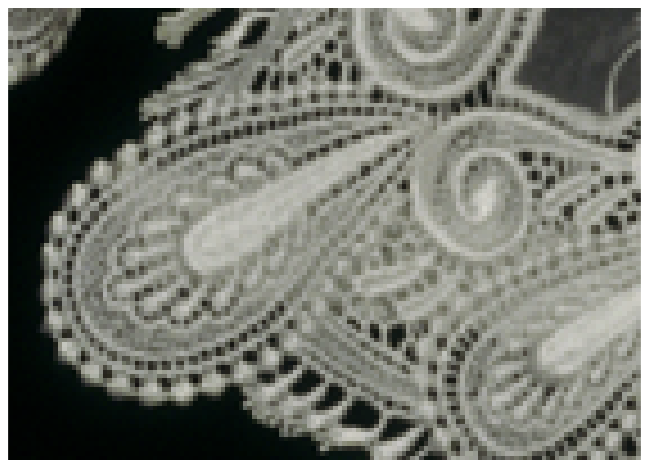}
         \caption{Original}
         \label{subfig:oriMcM}
     \end{subfigure}
	~
		\begin{subfigure}[b]{0.455\textwidth}
         \centering
         \includegraphics[width=1.0\textwidth]{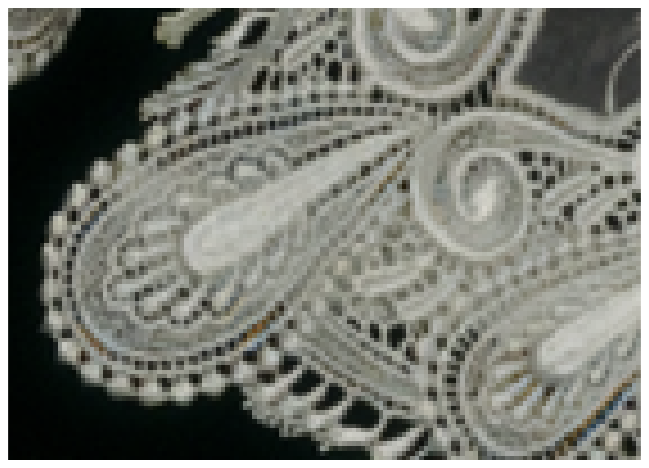}
         \caption{2-Stage}
         \label{subfig:2StageMcM}
     \end{subfigure}
	\\
	\vspace{0.5cm}
		\begin{subfigure}[b]{0.455\textwidth}
         \centering
         \includegraphics[width=1.0\textwidth]{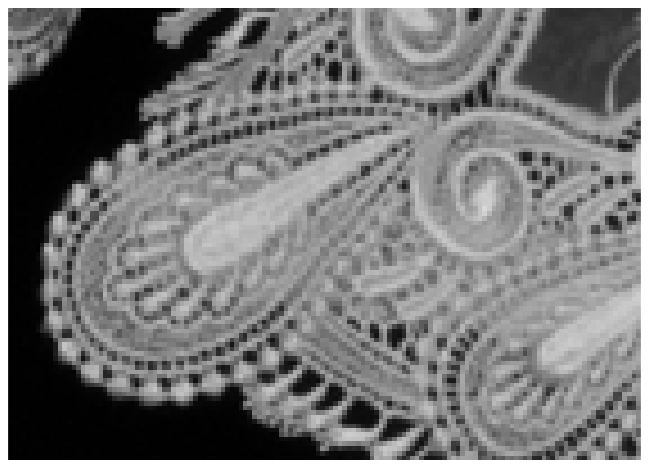}
         \caption{3-Stage}
         \label{subfig:3StageMcM}
     \end{subfigure}
	~
		\begin{subfigure}[b]{0.455\textwidth}
         \centering
         \includegraphics[width=1.0\textwidth]{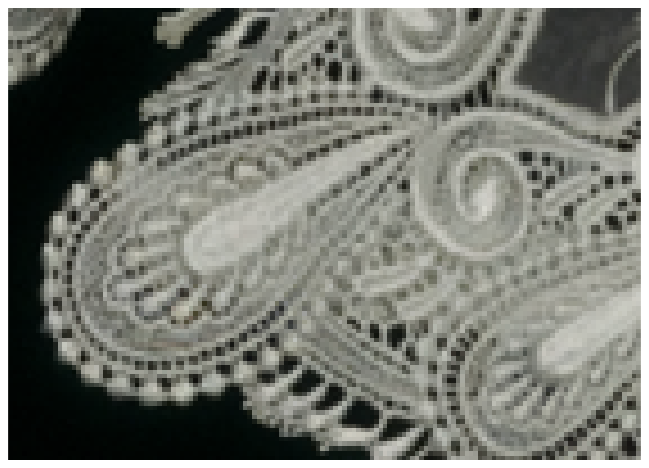}
         \caption{Proposed}
         \label{subfig:newMcM}
     \end{subfigure}
		\caption{A visual comparison example taken from the 8th image of McM. The 2-Stage network generates false colors, and the 3-Stage network miss-recovers the brightness. The proposed method reconstructs this region sharply and faithfully. Better viewed digitally.}
	\label{fig:McMBeauty}
\end{figure*}

Fig. \ref{subfig:oriMcM}-\ref{subfig:newMcM} compares the three models by a region taken from the 8th image of McM. The 2-Stage network suffers noticeable false colors, whereas the 3-Stage network and our method yield sharp recovery. However, the 3-Stage network exhibits an overall saturation bias in this region, making the object looks ``colder'' than the ground truth. 

\begin{figure*}[h]
	\centering
	\begin{subfigure}[b]{0.455\textwidth}
         \centering
         \includegraphics[width=1.0\textwidth]{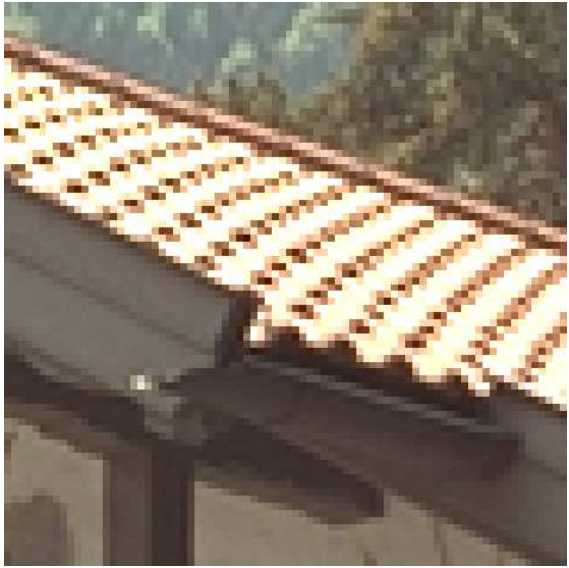}
         \caption{Original}
         \label{subfig:oriKodak}
     \end{subfigure}
	~
		\begin{subfigure}[b]{0.455\textwidth}
         \centering
         \includegraphics[width=1.0\textwidth]{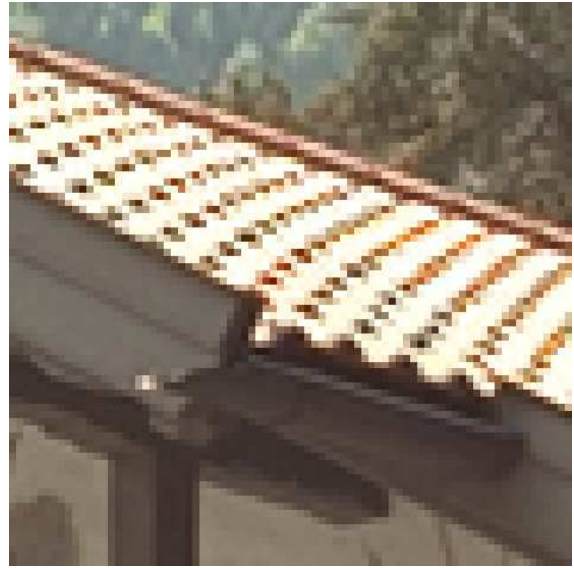}
         \caption{2-Stage}
         \label{subfig:2StageKodak}
     \end{subfigure}
	\\
		\begin{subfigure}[b]{0.455\textwidth}
         \centering
         \includegraphics[width=1.0\textwidth]{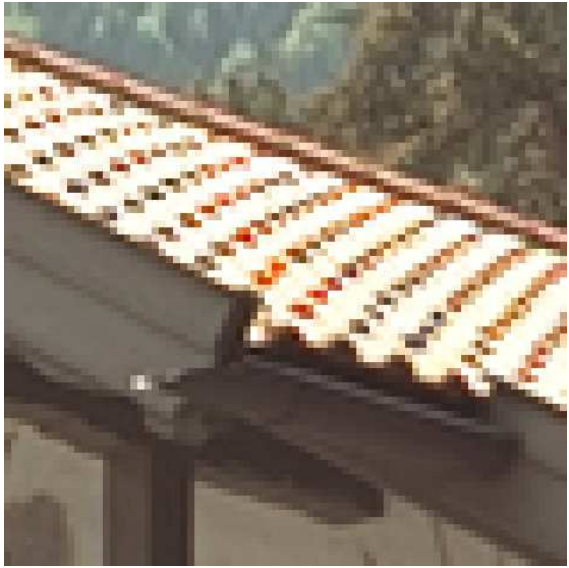}
         \caption{3-Stage}
         \label{subfig:3StageKodak}
     \end{subfigure}
	~
		\begin{subfigure}[b]{0.455\textwidth}
         \centering
         \includegraphics[width=1.0\textwidth]{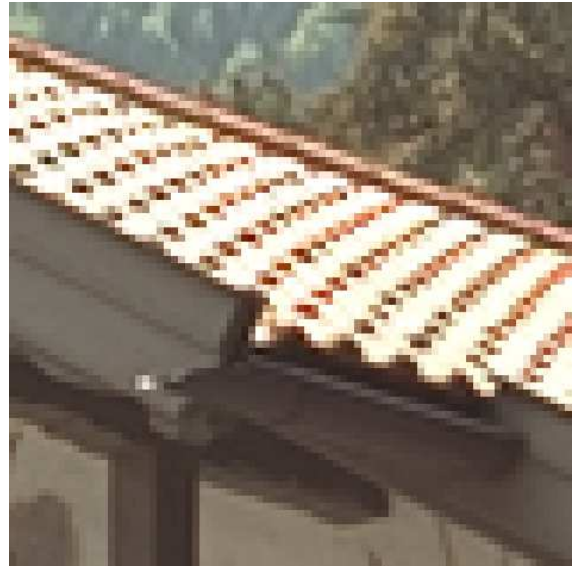}
         \caption{Proposed}
         \label{subfig:newKodak}
     \end{subfigure}
		\caption{A visual comparison example taken from the 24th image of Kodak. The 2-Stage network suffers false colors at textures in the left part of the region; the 3-Stage network suffers false colors at textures in the right part of the region. The proposed method has better visual quality in both parts. Better viewed digitally.}
	\label{fig:McMKodak}
\end{figure*}

Figures \ref{subfig:oriKodak}-\ref{subfig:newKodak} illustrate a region from the 24th image of Kodak. At the fine textures, the 2-Stage and 3-Stage networks both suffer the false color aliasing, whereas our result is visually close and faithful to the original image.

\section{Conclusion}
This paper proposes a novel de-Bayer technique based on light weight convolutional neural networks. Having observed that the recovery of the three channels has different complexity, we model them separately in parallel. By transforming the red and blue channel recovery to the green-red and green-blue difference recovery, the complexity of the recovery can be effectively reduced.  We have demonstrated that, this strategy improves the demosaicking efficiency without foregoing the accuracy. We have presented new approaches to progressively searching for small networks with low validation errors, leveraging skip connections and dilated convolutions. As each sub-network is of light weight, the total number of parameters and Flops are remarkably smaller than existing networks that have parallel multi-network structures. We have compared the accuracy and model complexity of our method to state-of-the-art CNN-based methods. The intensive comparison study has verified that our algorithm generalizes well to both conventional and recent test datasets. Compared with CNN-based methods of similar accuracy, our model is substantially smaller and compact. Compared to existing light weight models, our method achieves higher demosaicking quality on either easy or hard benchmarks.



\begin{thebibliography}{10}
\providecommand{\url}[1]{#1}
\csname url@samestyle\endcsname
\providecommand{\newblock}{\relax}
\providecommand{\bibinfo}[2]{#2}
\providecommand{\BIBentrySTDinterwordspacing}{\spaceskip=0pt\relax}
\providecommand{\BIBentryALTinterwordstretchfactor}{4}
\providecommand{\BIBentryALTinterwordspacing}{\spaceskip=\fontdimen2\font plus
\BIBentryALTinterwordstretchfactor\fontdimen3\font minus
  \fontdimen4\font\relax}
\providecommand{\BIBforeignlanguage}[2]{{%
\expandafter\ifx\csname l@#1\endcsname\relax
\typeout{** WARNING: IEEEtran.bst: No hyphenation pattern has been}%
\typeout{** loaded for the language `#1'. Using the pattern for}%
\typeout{** the default language instead.}%
\else
\language=\csname l@#1\endcsname
\fi
#2}}
\providecommand{\BIBdecl}{\relax}
\BIBdecl

\bibitem{szeliski2010}
R.~Szeliski, \emph{Computer vision: algorithms and applications}.\hskip 1em
  plus 0.5em minus 0.4em\relax Springer Science; Business Media, 2010.

\bibitem{Buckler17}
M.~Buckler, S.~Jayasuriya, and A.~Sampson, ``Reconfiguring the imaging pipeline
  for computer vision,'' \emph{IEEE International Conference on Computer
  Vision}, pp. 975--984, 2017.

\bibitem{Cok87}
D.~R. Cok, ``Signal processing method and apparatus for sampled image
  signals,'' 1987, [US Patent 4 630 307].

\bibitem{HA96}
J.~F. {Hamilton Jr.} and J.~E. Adams, ``Adaptive color plane interpolation in
  single sensor color electronic camera,'' 1997, [US Patent 5,629,734].

\bibitem{Niu2019}
Y.~Niu, J.~Ouyang, W.~Zuo, and F.~Wang, ``Low cost edge sensing for high
  quality demosaicking,'' \emph{IEEE TIP}, vol.~28, no.~5, 2019.

\bibitem{LDINAT11}
L.~Zhang, X.~Wu, A.~Buades, and X.~Li, ``Color demosaicking by local
  directional interpolation and nonlocal adaptive thresholding,'' \emph{Journal
  of Electronic imaging}, vol.~20, no.~2, p. 023016, 2011.

\bibitem{MSR2014}
D.~Khashabi, S.~Nowozin, J.~Jancsary, and A.~Fitzgibbon, ``Joint demosaicing
  and denoising via learned nonparametric random fields,'' \emph{IEEE Trans. on
  Image Proc.}, vol.~23, no.~12, pp. 4968--4981, 2014.

\bibitem{LDSR14}
M.~Rossi and G.~Calvagno, ``Luminance driven sparse representation based
  demosaicking,'' in \emph{IEEE International Conference on Image Processing},
  2014, pp. 1788--1792.

\bibitem{DDR16}
J.~Wu, R.~Timofte, and L.~V. Gool., ``Demosaicing based on directional
  difference regression and efficient regression priors,'' \emph{IEEE
  Transactions on Image Processing}, vol.~25, no.~8, pp. 3862--3874., 2016.

\bibitem{MLRI16}
D.~Kiku, Y.~Monno, M.~Tanaka, and M.~Okutomi, ``Beyond color difference:
  residual interpolation for color image demosaicking,'' \emph{IEEE
  Transactions on Image Processing}, vol.~25, no.~3, pp. 1288--1300, 2016.

\bibitem{HQLI04}
H.~S. Malvar, L.~He, and R.~Cutler, ``High-quality linear interpolation for
  demosaicing of bayer-patterned color images,'' in \emph{Proc. IEEE ICASSP},
  2004, pp. iii--485--8, [Matlab build-in function \textbf{demosaic()}].

\bibitem{MNN14}
Y.-Q. Wang, ``A multilayer neural network for image demosaicking,'' in
  \emph{Image Processing, 2014 IEEE International Conference on}, 2014, pp.
  1852--1856.

\bibitem{DRL17}
R.~Tan, K.~Zhang, W.~Zuo, and L.~Zhang, ``Color image demosaicking via deep
  residual learning,'' in \emph{2017 IEEE International Conference on
  Multimedia and Expo}, 2017, pp. 793--798.

\bibitem{DJ16}
M.~Gharbi, G.~Chaurasia, S.~Paris, and F.~Durand, ``Deep joint demosaicking and
  denoising,'' \emph{ACM Transactions on Graphics}, vol.~35, no.~6, p. 191,
  2016.

\bibitem{CAS18}
F.~Kokkinos and S.~Lefkimmiatis, ``Deep image demosaicking using a cascade of
  convolutional residual denoising networks,'' in \emph{European Conference on
  Computer Vision}, 2018.

\bibitem{BurstCVPR19}
------, ``Iterative residual cnns for burst photography applications,'' in
  \emph{IEEE Conference on Computer Vision and Pattern Recognition}, 2019, pp.
  5929--5938.

\bibitem{BurstICCV19}
T.~Ehret, A.~Davy, P.~Arias, and G.~Facciolo, ``Joint demosaicking and
  denoising by fine-tuning of bursts of raw images,'' in \emph{IEEE
  International Conference on Computer Vision}, 2019, pp. 8868--8877.

\bibitem{MDFCN18}
D.~S. Tan, W.-Y. Chen, and K.-L. Hua, ``Deepdemosaicking: Adaptive image
  demosaicking via multiple deep fully convolutional networks,'' \emph{IEEE
  Transactions on Image Processing}, vol.~27, no.~5, pp. 2408--2419, 2018.

\bibitem{ICASSP19}
T.~Yamaguchi and M.~Ikehara, ``Image demosaiking via chrominance images with
  parallel convolutional neural networks,'' in \emph{International Conference
  on on Acoustics, Speech and Signal Processing}, 2019, pp. 1702--1706.

\bibitem{huang2018}
T.~Huang, F.-F. Wu, W.~Dong, G.~Shi, and X.~Li, ``Lightweight deep residue
  learning for joint color image demosaicking and denoising,'' in
  \emph{International Conference on Pattern Recognition}, 2018, pp. 127--132.

\bibitem{Shuffle18}
N.~Ma, X.~Zhang, H.-T. Zheng, and J.~Sun, ``Shufflenet v2: Practical guidelines
  for efficient cnn architecture design,'' in \emph{The European Conference on
  Computer Vision}, 2018, pp. 116--131.

\bibitem{he2016deep}
K.~He, X.~Zhang, S.~Ren, and J.~Sun, ``Deep residual learning for image
  recognition,'' in \emph{IEEE CVPR}, 2016, pp. 770--778.

\bibitem{autoencoder14}
G.~Alain and Y.~Bengio, ``What regularized auto-encoders learn from the
  data-generating distribution,'' \emph{J. Mach. Learn. Res.}, vol.~15, no.~1,
  pp. 3563--3593, 2014.

\bibitem{Waterloo17}
K.~Ma, Z.~Duanmu, Q.~Wu, Z.~Wang, H.~Yong, H.~Li, and L.~Zhang, ``{Waterloo
  Exploration Database}: New challenges for image quality assessment models,''
  \emph{IEEE Transactions on Image Processing}, vol.~26, no.~2, pp. 1004--1016,
  2017.

\bibitem{kingma2015}
D.~P. Kingma and J.~Ba, ``Adam: A method for stochastic optimization,'' in
  \emph{International Conference on Learning Representations}, vol.~5, 2015.

\bibitem{vedaldi15matconvnet}
A.~Vedaldi and K.~Lenc, ``Matconvnet -- convolutional neural networks for
  matlab,'' in \emph{{ACM} Int. Conf. on Multimedia}, 2015.

\bibitem{Kodak}
``Low resolution kodak image dataset,'' \url{http://r0k.us/graphics/kodak/},
  1991.

\bibitem{JDD2018}
B.~Henz, E.~S.~L. Gastal, and M.~M. Oliveira, ``Deep joint design of color
  filter arrays and demosaicing,'' \emph{Eurographics}, vol.~37, no.~2, pp.
  389--399, 2018.

\bibitem{3Stage18}
K.~Cui, Z.~Jin, and E.~Steinbach, ``Color image demosaicking using a 3-stage
  convolutional neural network structure,'' in \emph{IEEE International
  Conference on Image Processing}, 2018, pp. 2177--2181.

\bibitem{SuperTex}
D.~Dai, R.~Timofte, and L.~{Van Gool}, ``Jointly optimized regressors for image
  super-resolution,'' in \emph{Eurographics}, 2015.

\end{thebibliography}

\end{document}